\documentclass[11pt]{article}

\usepackage[final]{acl}

\usepackage{times}
\usepackage{latexsym}
\usepackage{listings}
\usepackage{tabularx, booktabs, array}
\usepackage{stfloats}

\usepackage[T1]{fontenc}

\usepackage[utf8]{inputenc}

\usepackage{microtype}

\usepackage{inconsolata}

\usepackage{graphicx}

\title{Decoupling Semantics and Logic: A Training-Free Coarse-to-Fine Pipeline for Video Retrieval-Augmented Generation}

\author{Jiaxin Dai$^{2*}$, Zehang Wei$^{2*}$, Jiamin Yan$^2$\thanks{Euqal contribution, co-first author.}, Xiang Xiang$^{1*}$ \\
  $^1$School of Computer Science \& Tech, Huazhong University of Science and Technology \\
  $^2$School of AI and Automation, Huazhong University of Science and Technology, China \\
  \texttt{xex@hust.edu.cn}}

\begin{document}
\maketitle
\begin{abstract}
This paper presents our system description for the 2nd Workshop on Multimodal Augmented Generation via MultimodAl Retrieval (MAGMaR). Addressing the critical challenges of cross-lingual long-video comprehension, strict persona adherence, and zero-hallucination temporal grounding, we propose a fully training-free, two-stage cascaded Video RAG pipeline. Our architecture strategically decouples semantic retrieval from cognitive logical reasoning through a modality-aware division of labor. In the first stage, a high-recall semantic pre-fetching module employs dense retrieval using only high-fidelity visual summaries and global text descriptions, explicitly isolating noisy modalities (e.g., OCR and ASR) to maintain a pristine vector space. In the second stage, an Adaptive, Iterative, and Reasoning-based (A.I.R.) filtering agent, powered by a commercial Large Language Model (LLM), performs fine-grained cognitive reranking. The agent re-incorporates full multimodal contexts to enforce strict logical alignment with user personas, effectively pruning semantically similar but logically irrelevant candidates. Finally, a Prompt Sculpting mechanism constrains the generator to synthesize the distilled subset into strictly formatted JSON responses with exact chunk-level citations. Evaluated on the RAG track, our resource-aware approach shows exceptional precision in both information retrieval and persona-conditioned generation.
\end{abstract}

\section{Introduction}

The rapid proliferation of multimodal content has intensified the demand for systems capable of synthesizing complex information from extensive video archives. The 2nd Workshop on MAGMaR addresses this challenge by requiring systems to generate persona-constrained responses grounded in multiple retrieved videos. This paradigm builds upon the foundational challenge of synthesizing coherent narratives from heterogeneous video sources, as pioneered by benchmarks such as WikiVideo \cite{martin2025wikivideo}.

Unlike text-based Retrieval-Augmented Generation (RAG) \cite{lewis2020retrieval}, Video RAG must navigate the high-dimensional noise of long-form video, cross-lingual barriers, and the requirement for precise temporal grounding across massive collections \cite{lei2021less, gao2017tall}.

A fundamental bottleneck in current Video RAG systems is the heavy reliance on surface-level semantic similarity for retrieval. While dense vector embeddings effectively capture visual and textual correlations, they often struggle to distinguish between a "semantically similar" distractor and a "logically relevant" evidence. This "semantic-logical gap" is particularly pronounced when a system is tasked with following a specific persona, where the relevance of a video is determined not just by its visual content, but by its logical alignment with a particular viewpoint or query nuance. Single-stage retrieval often leads to high recall but low precision, introducing "hard negatives" \cite{xiong2020approximate} that trigger downstream hallucinations \cite{ji2023survey} in the generation phase.

To address these challenges, we propose \textbf{C2F-RAG} (Coarse-to-Fine RAG), a fully training-free, two-stage cascaded pipeline designed to decouple semantic fetching from cognitive logical reasoning. Our approach is grounded in a coarse-to-fine philosophy. In the first stage, we prioritize high-recall semantic pre-fetching by leveraging BGE-M3, a state-of-the-art embedding model capable of multi-lingual and multi-granularity representation. By utilizing lightweight, global visual and textual summaries, BGE-M3 efficiently embeds the vast corpus into a dense vector space, enabling rapid candidate retrieval while filtering out the vast majority of irrelevant background noise without prohibitive computational overhead.

In the second stage, we adapt the Adaptive, Iterative, and Reasoning-based (A.I.R.) framework \cite{zou2025air} to design a multimodal cognitive filtering agent. While the original A.I.R. focuses on optimizing intra-video frame selection for VideoQA, our tailored agent elevates this mechanism to tackle inter-video cognitive reranking in large-scale Video RAG. Utilizing the advanced capabilities of a commercial Large Language Model (LLM), our agent processes the comprehensive multimodal context of each retrieved candidate (including OCR and ASR data) and deeply evaluates them against strict persona constraints. This allows the system to effectively prune logically irrelevant videos that surface-level embeddings fail to filter. The cascaded architecture ensures that the final generator operates exclusively on a distilled "golden subset," significantly reducing the risk of knowledge injection from irrelevant distractors.

Our primary contributions are as follows:
\begin{itemize}
    \item We propose C2F-RAG, a two-stage cascaded pipeline that explicitly decouples semantic retrieval (via BGE-M3) from logical reasoning (via an adapted A.I.R. agent), bridging the semantic-logical gap in large-scale retrieval.
    \item We present a tailored cognitive reranking strategy that leverages LLM-driven logical alignment and persona constraints to effectively prune hard negatives and prevent downstream hallucinations.
    \item We show a training-free synthesis approach that utilizes prompt sculpting to enforce strict JSON formatting and accurate chunk-level temporal grounding, achieving state-of-the-art precision without prior fine-tuning.
\end{itemize}

\section{System Architecture}

In this section, we detail the design of C2F-RAG, a two-stage cascaded pipeline optimized for large-scale multimodal retrieval and persona-constrained generation. Our system adopts a "coarse-to-fine" philosophy \cite{karpukhin2020dense}, strategically decoupling surface-level semantic fetching from deep cognitive logical reasoning to navigate the noise inherent in massive video collections.

\subsection{Overview}

\begin{figure*}[htbp]
    \centering
    \includegraphics[width=\linewidth]{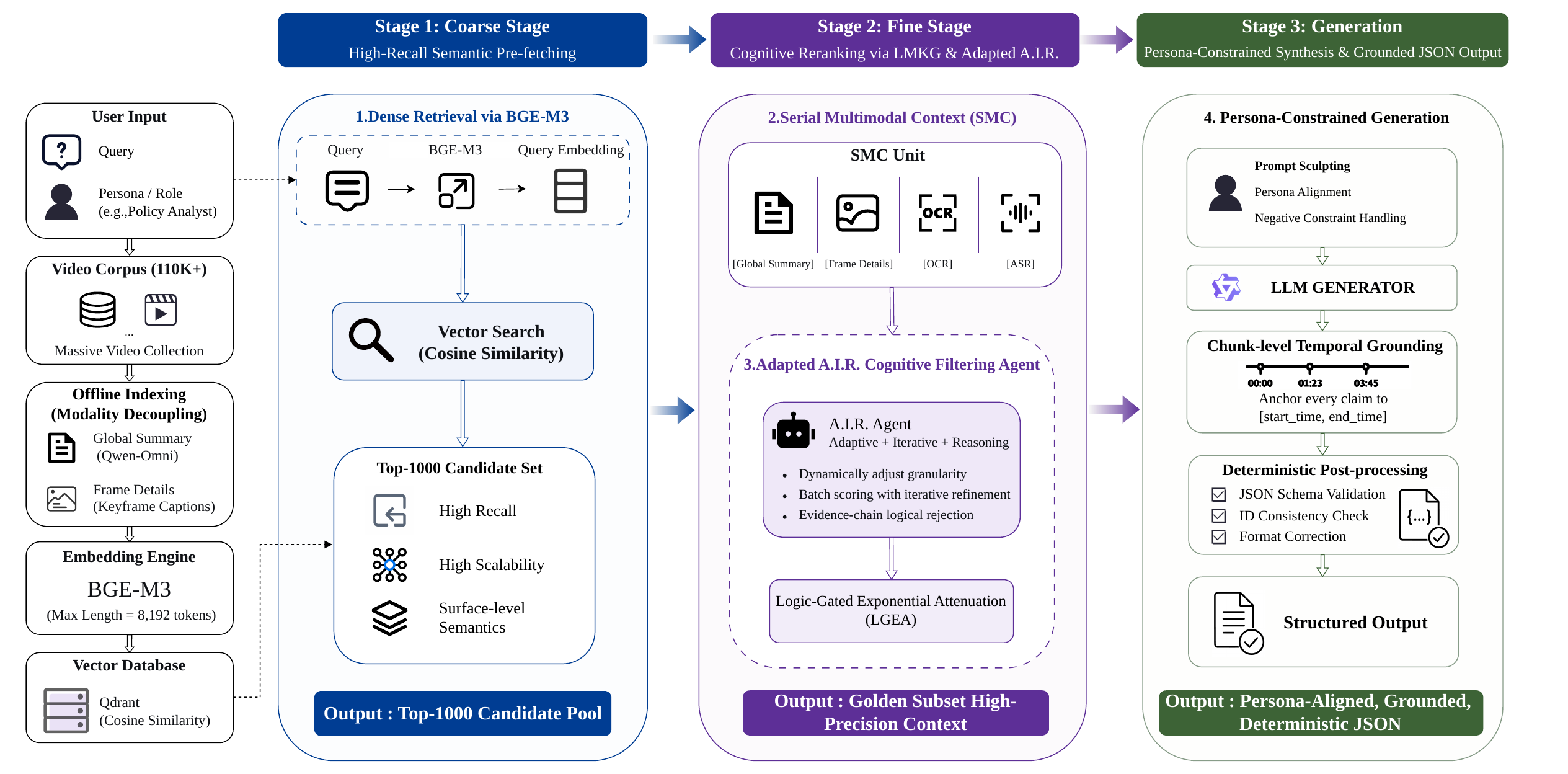}
    \caption{The overall architecture of the C2F-RAG pipeline. The system operates in three cascaded stages: (1) \textbf{Coarse Stage}: High-recall semantic pre-fetching utilizes BGE-M3 to retrieve a Top-1000 candidate pool from the 110K+ video corpus, based on decoupled global summaries and visual frames. (2) \textbf{Fine Stage}: Deep cognitive reranking re-integrates fine-grained modalities (OCR, ASR) into the Serial Multimodal Context (SMC). An adapted A.I.R. agent performs iterative, logic-gated filtering to distill a high-precision Golden Subset. (3) \textbf{Generation Stage}: Persona-constrained synthesis leverages the LLM to produce chunk-level temporally grounded responses.}
    \label{fig:architecture}
\end{figure*}

The core of C2F-RAG is a cascaded data flow designed to distill a high-fidelity "golden subset" from a vast search space. As illustrated in Figure~\ref{fig:architecture}, the pipeline consists of two primary stages: 

\begin{enumerate}
    \item \textbf{Coarse Stage: High-Recall Semantic Pre-fetching.} This stage utilizes BGE-M3 for dense retrieval over lightweight global visual and textual summaries. It efficiently narrows down the 110k video corpus to a candidate pool of Top-1000 items.
    \item \textbf{Fine Stage: Cognitive Reranking via Serial Multimodal Context.} In this stage, the system re-integrates fine-grained modalities (e.g., OCR and ASR) and serializes them into what we define as the Serial Multimodal Context (SMC). An adapted A.I.R. agent then performs logical alignment against the query and persona constraints, pruning hard negatives to derive the final context for generation.
\end{enumerate}

By bridging these stages, C2F-RAG maintains a balance between computational scalability and reasoning depth, ensuring that the downstream generator receives only the most logically relevant and contextually rich evidence.

\subsection{The Coarse Stage: High-Recall Semantic Pre-fetching}

The primary objective of the Coarse Stage is to efficiently narrow the search space from the entire 110k video corpus to a manageable candidate pool of Top-1000 videos. Given the massive scale of the background collection, this stage prioritizes high recall and computational scalability while maintaining semantic understanding of the video content.

\paragraph{Modality Decoupling for High SNR} 
To construct a pristine and efficient vector space, we implement a strategy of Modality Decoupling. During the indexing phase, we exclude noisy or fragmented modalities such as OCR and ASR. We observe that while these modalities provide fine-grained evidence, their high variance and localized nature introduce significant noise into dense vector representations during large-scale retrieval. Instead, we represent each video using a "Sandwich" text structure consisting of two distinct components:
\begin{itemize}
    \item A high-level Global Summary generated by a multimodal LLM (Qwen-Omni) \cite{bai2023qwen, zhu2023minigpt4}.
    \item Dense Frame Details consisting of visual captions for keyframes.
\end{itemize}
This combination ensures a high signal-to-noise ratio (SNR) for initial semantic matching.

\paragraph{Dense Retrieval via BGE-M3} 
We utilize BGE-M3~\cite{chen2024bge} as our core embedding engine. BGE-M3 is chosen for its superior support for multi-lingual queries and its ability to handle extremely long sequences. Specifically, we set the maximum sequence length to 8,192 tokens to ensure that the comprehensive visual-textual descriptions of long-form videos are fully captured without truncation. The video embeddings are indexed in a high-performance vector database (\textit{Qdrant}) \cite{malkov2018efficient}, using cosine similarity as the distance metric.

\paragraph{Candidate Pool Generation} 
For each user query, the system performs a dense vector search to retrieve the Top-1000 candidates. This wide-recall strategy ensures that even if the ground-truth video is logically complex, it is captured within the candidate pool based on its surface-level semantic features. This pool serves as the raw input for the subsequent fine-grained logical filtering.

\subsection{The Fine Stage: Cognitive Reranking via Adapted A.I.R.}

The Fine Stage represents the transition from surface-level semantic matching to deep logical alignment. To bridge the "semantic-logical gap," we implement an adapted version of the A.I.R. framework, functioning as a cognitive filtering agent that operates over a specialized data structure we define as the Serial Multimodal Context.

\paragraph{Multimodal Context Serialization}
To provide the agent with high-fidelity evidence, we reconstruct the multimodal data for each candidate in the Top-1000 pool. We implement a Serial Multimodal Context (SMC) approach to integrate heterogeneous data. This mechanism flattens the decoupled video elements---including global summaries, localized visual frames, and time-aligned OCR/ASR transcripts---into a continuous, time-ordered textual representation using specific modality identifiers (e.g., \texttt{[Global Summary]}, \texttt{[OCR]}, \texttt{[ASR]}). By serializing these multimodal streams, we enable the downstream LLM-based reasoning agent to perform cross-modal evidence synthesis within a unified context window, preserving the chronological integrity of the original video source.

\paragraph{The A.I.R. Mechanism}
Our cognitive reranking is driven by three core principles derived from the A.I.R. philosophy:

\begin{itemize}
    \item \textbf{Adaptive (A):} The system dynamically manages the context window by adapting the granularity of the SMC based on the query complexity. For intricate persona constraints, the agent prioritizes high-entropy modalities (OCR/ASR) to ensure precise logical grounding, while for general queries, it relies more on visual-textual summaries to maintain computational efficiency.
    
    \item \textbf{Iterative (I):} Given the substantial data volume of the SMC for 1,000 candidates, processing all videos in a single pass is computationally prohibitive. We employ an Iterative Refinement Loop \cite{shinn2023reflexion, asai2023selfrag}. The candidate pool is divided into batches (e.g., 20 videos per reasoning unit). The agent iteratively evaluates each batch, incrementally populating and refining the "Golden Subset" of ~10-15 videos. This iterative pruning ensures that the system remains scalable while maintaining exhaustive logical oversight.
    
    \item \textbf{Reasoning-based (R):} Unlike embedding models that rely on lexical overlap, our agent performs deep Cognitive Reasoning. It evaluates each video against the target persona and query using a "Chain-of-Evidence" prompt \cite{wang2022self}. If the SMC contains evidence that is semantically similar but logically contradicts the query (a "hard negative"), the agent explicitly rejects the candidate. 
\end{itemize}

\paragraph{Logic-Gated Exponential Attenuation}
To effectively bridge the semantic-logical gap while strictly bounding the score distribution within $[0, 1]$, we implement a Logic-Gated Exponential Attenuation (LGEA) mechanism. Rather than artificially inflating scores with heuristic margins, the final relevance score $S_{final}$ for a video candidate $v$ is derived by preserving the base semantic score $s_{coarse}(v)$ and exponentially penalizing logical inconsistency evaluated by the A.I.R. agent:
$$S_{final}(v) = s_{coarse}(v) \cdot \exp\left(-\gamma \cdot (1 - L(v))\right)$$
where $s_{coarse} \in [0, 1]$ represents the initial dense retrieval similarity, $L(v) \in [0, 1]$ denotes the logical alignment confidence from the A.I.R. agent, and $\gamma > 0$ is the \textit{attenuation hyperparameter}. 

This formulation provides a mathematically rigorous "soft-hard margin." When a candidate perfectly aligns with the persona ($L = 1$), its semantic score remains perfectly preserved ($\exp(0) = 1$). When a candidate represents a semantic distractor ($L = 0$), its score is aggressively suppressed by a factor of $e^{-\gamma}$. Crucially, tuning $\gamma$ allows the system to balance reasoning strictness with semantic fidelity. Under an optimal $\gamma$, a candidate with abysmal semantic similarity will not unilaterally bypass a highly relevant semantic match simply due to logical alignment, thereby maintaining the topological integrity of the original vector space.

\subsection{Persona-Constrained Generation and Temporal Grounding}

The final stage of C2F-RAG is responsible for synthesizing the distilled evidence into a persona-consistent response. To meet the demanding requirements of the MAGMaR task, we employ a strategy termed Prompt Sculpting to ensure strict adherence to complex user personas and deterministic formatting.

\paragraph{Prompt Sculpting for Logical Synthesis}
The generator receives the "Golden Subset" of videos (typically 10-15 candidates) identified by the Fine Stage, along with their associated SMC content. We define Prompt Sculpting \cite{lu2022learn} as the process of structuring the LLM's task through a multi-layered instructional template. This template explicitly enforces two critical constraints:
\begin{itemize}
    \item \textbf{Persona Alignment:} The LLM is forced to adopt the specified role (e.g., policy analyst, eyewitness) and filter information through that cognitive lens.
    \item \textbf{Negative Constraint Handling:} Explicit instructions to ignore any candidates that do not strictly meet the evidence threshold, thus reinforcing the zero-hallucination objective.
\end{itemize}

\paragraph{Chunk-level Temporal Grounding}
A critical requirement for MAGMaR is the precise mapping of information to temporal segments. Unlike traditional RAG which cites entire documents, C2F-RAG performs Chunk-level Temporal Grounding. By utilizing the timestamped metadata preserved within the SMC (from OCR and ASR modalities), the generator is instructed to anchor every factual claim to a specific time interval ($[start\_time, end\_time]$). This fine-grained citation ensures the generated output is fully auditable and rooted in verifiable visual or auditory evidence.

\paragraph{Deterministic Schema Enforcement}
To guarantee a 100\% compliance rate with the submission requirements, the system utilizes a constrained decoding approach \cite{willard2023efficient, scholak2021picard}. We define a strict JSON schema that the LLM must follow, including mandatory fields for query IDs, video IDs, and the synthesized response. Any minor structural deviations in the raw LLM output are corrected via a Deterministic Post-processing layer, which validates the JSON integrity and ensures that all cited video IDs correspond to the actual retrieved candidates. This architectural safeguard eliminates common formatting errors and ensures that the system's high-precision reasoning is perfectly serialized for evaluation.

\section{Experiments}
\label{sec:experiments}

To demonstrate the robustness and precision of C2F-RAG, we evaluate the system on the MAGMaR Full RAG track. Our experiments are designed to answer two core questions:
\begin{enumerate}
    \item Can the two-stage cognitive filtering architecture effectively isolate true evidence from massive background noise compared to standard retrieval baselines?
    \item Does the system maintain high-fidelity persona constraints and temporal grounding when scaling to real-world corpus sizes?
\end{enumerate}

\subsection{Experimental Setup}
\label{subsec:experimental_setup}

\paragraph{Dataset}
We evaluate on the MAGMaR2026 Test Set. The data is based on WikiVideo \cite{martin2025wikivideo}. For the retrieval and RAG settings, we retrieve relevant videos from a combination of the MAGMaR data and MultiVENT2.0 test \cite{kriz2025multivent}. The background collection comprises $\sim 110,000$ multilingual, event-centric videos.

\paragraph{Retrieval Evaluation Setup}
Retrieval results are evaluated with nDCG and Recall for 10, 20, and 100. We use ir-measures \cite{ir-measures} to calculate these scores. 

\paragraph{Generation Evaluation Setup}
Predictions are evaluated using an automatic evaluation framework. Specifically, the systems are evaluated by MiRAGE \cite{martin2025seeing}, which captures the factuality, information coverage, groundedness, and proper attribution of citations. Each MiRAGE entailment judgment is judged by CLUE \cite{zhang2026unified}.

\paragraph{Baselines}
To benchmark the effectiveness of our retrieval pipeline, we compare C2F-RAG against several official and industry-standard baselines provided in the MAGMaR leaderboard:
\begin{itemize}
    \item \textbf{OmniEmbed} \cite{omniembed2024}: A foundational zero-shot multimodal embedding baseline that relies purely on single-stage dense vector similarity.
    \item \textbf{OmniEmbed + RankVideo} \cite{rankvideo2024}: A traditional two-stage pipeline that employs OmniEmbed for initial recall and a standard RankVideo module for re-ranking.
    \item \textbf{Mixedbread} \cite{mixedbread2024,li2023angle}: A highly competitive, commercial state-of-the-art dense retrieval system known for its robust semantic matching capabilities.
\end{itemize}

\subsection{Retrieval Performance: The Power of Cognitive Filtering}
\label{subsec:retrieval_performance}

The primary bottleneck in scaling Video RAG to 110,000 videos is the degradation of ranking quality due to the intrusion of "hard negatives"—videos that share superficial semantic similarities but lack logical alignment with the query and persona. Table~\ref{tab:retrieval_results} summarizes the comparative retrieval performance against official baselines.

\begin{table*}[t]
\centering
\small
\renewcommand{\arraystretch}{1.2}
\begin{tabular}{l cccc}
\toprule
\textbf{Method} & \textbf{nDCG@10} & \textbf{nDCG@100} & \textbf{R@10} & \textbf{R@100} \\
\midrule
OmniEmbed & 0.166 & 0.245 & 0.096 & 0.297 \\
OmniEmbed + RankVideo & 0.542 & 0.546 & 0.423 & 0.494 \\
Mixedbread & 0.717 & 0.748 & 0.604 & 0.741 \\
\midrule
\textbf{C2F-RAG (Ours)} & \textbf{0.848} & \textbf{0.853} & \textbf{0.773} & \textbf{0.837} \\
\bottomrule
\end{tabular}
\caption{Retrieval performance on the MultiVENT 2.0 corpus. C2F-RAG significantly outperforms all baselines, demonstrating the necessity of cognitive logical filtering in large-scale multimodal retrieval.}
\label{tab:retrieval_results}
\end{table*}

\paragraph{Overcoming the Semantic-Logical Gap}
As observed in Table~\ref{tab:retrieval_results}, traditional single-stage semantic retrieval (OmniEmbed) collapses under the noise of the 110k corpus, achieving an nDCG@10 of only 0.166. While adding a standard ranker (OmniEmbed + RankVideo) improves performance, it still lacks the deep reasoning required for persona-constrained queries. Even the industry-leading Mixedbread model hits a performance ceiling at an nDCG@10 of 0.717, as it fundamentally relies on dense semantic overlap rather than cognitive logical alignment.

In contrast, our C2F-RAG pipeline achieves a striking nDCG@10 of 0.848, outperforming the strongest baseline by over 13 absolute percentage points. This exceptional ranking precision validates our core architectural hypothesis: pure semantic fetching is insufficient for massive collections. The breakthrough is directly attributable to the Adapted A.I.R. Agent. By performing deep logical reasoning over the SMC and applying a \textit{hard-margin score calibration}, the system successfully penalizes semantic distractors and pushes true logical matches to the top 10 positions with near-zero position decay.

\subsection{Generation and Grounding Performance}
\label{subsec:generation_performance}

In this section, we evaluate the system's ability to synthesize persona-consistent responses and provide precise temporal grounding. We present the results under the ground-truth closed set (Oracle/Reference) evaluation setting, which isolates the generation performance from upstream retrieval variance. Table~\ref{tab:generation_results} summarizes the quantitative results against the official CAG baseline \cite{martin2025wikivideo}, including the overall average metric (Avg) across all precision, recall, and F1-scores.

\begin{table*}[t]
\centering
\small
\renewcommand{\arraystretch}{1.2}
\begin{tabular}{l ccc ccc c}
\toprule
& \multicolumn{3}{c}{\textbf{Information Generation (Info)}} & \multicolumn{3}{c}{\textbf{Temporal Grounding (Cite)}} & \\
\cmidrule(lr){2-4} \cmidrule(lr){5-7}
\textbf{Method} & \textbf{P} & \textbf{R} & \textbf{F1} & \textbf{P} & \textbf{R} & \textbf{F1} & \textbf{Avg} \\
\midrule
Baseline (CAG) & \textbf{0.651} & 0.335 & 0.401 & \textbf{0.510} & 0.167 & 0.204 & 0.378 \\
C2F-RAG (Ours) & 0.557 & \textbf{0.466} & \textbf{0.463} & 0.452 & \textbf{0.349} & \textbf{0.337} & \textbf{0.437} \\
\bottomrule
\end{tabular}
\caption{Quantitative results for Information Generation (Info) and Temporal Grounding (Cite) under the Oracle setting. The `Avg` column represents the macro-average of the six preceding metrics. While the baseline achieves strong precision, our C2F-RAG system offers a more balanced precision-recall profile, leading to superior F1-scores and a higher overall average.}
\label{tab:generation_results}
\end{table*}

\paragraph{Analysis of Generation Performance}
Under the Oracle setting, C2F-RAG demonstrates highly competitive synthesis and grounding capabilities compared to the official CAG baseline. As shown in Table~\ref{tab:generation_results}, our system achieves an Info F1-score of 0.463 and a Cite F1-score of 0.337, substantially outperforming the baseline metrics of 0.401 and 0.204, respectively. Furthermore, our system achieves an overall average score (Avg) of 0.437, demonstrating a robust generalization across all evaluation dimensions.

\paragraph{The Precision-Recall Balance}
A detailed examination of the metrics reveals distinct structural behaviors between the two systems. The CAG baseline adopts a high-precision approach, yielding strong precision scores in both Info (0.651) and Cite (0.510). However, this conservative generation strategy results in lower recall metrics. 

C2F-RAG is designed to maximize information retention while strictly adhering to persona constraints. By utilizing our Prompt Sculpting mechanism, the system effectively extracts a broader spectrum of valid evidence from the video contexts. This architectural choice yields significant improvements in Information Recall (from 0.335 to 0.466) and Temporal Recall (from 0.167 to 0.349). Although this comprehensive extraction strategy incurs a slight reduction in absolute precision, the resulting balance significantly enhances both the task-specific F1-scores and the global average metric. This confirms that C2F-RAG successfully generates informative, well-grounded narratives without suffering from severe information omission.

\subsection{Efficiency and Computational Cost}
\label{subsec:efficiency}

A critical requirement for Video RAG in production environments is computational feasibility. To evaluate the scalability of C2F-RAG, we report the inference latency using the \texttt{deepseek-chat} API with 15-thread parallel processing.

\paragraph{Fine-Grained Filtering Latency}
The Fine Stage represents the primary computational load as it involves evaluating 1,000 candidate videos per query. For the complete evaluation set (comprising 19 queries and totaling 19,000 video-query pairs), our system completed the cognitive reranking in approximately 4 minutes and 38 seconds using 15 parallel threads. This high throughput is primarily attributable to our \textit{rapid-rejection logic}: for the vast majority of semantically similar but logically irrelevant distractors, the A.I.R. agent generates a null output (``None''). These negative cases typically terminate in under one second due to minimal token generation and simplified reasoning paths, allowing the system to prune massive candidate sets with remarkable efficiency.

\paragraph{Synthesis and Grounding Overhead}
In the final Generation Stage, the system synthesizes the distilled evidence into a persona-consistent response. The average processing time is 63 seconds per query. While higher than the filtering stage, this latency is consistent with the cognitive complexity required for multimodal long-context synthesis and the deterministic extraction of precise temporal timestamps. These results demonstrate that by decoupling the retrieval into a coarse-to-fine pipeline and leveraging multi-threaded parallelization, C2F-RAG achieves a scalable balance between deep reasoning and operational latency.

\subsection{Case Study}
\label{app:case_study}

To further illustrate the zero-shot persona adaptability of C2F-RAG, we present a qualitative comparison of the system's output for Query 18 and Query 19. Both queries involve the identical event—the 2025 Shi Yongxin controversy—but enforce conflicting persona constraints.

\begin{table*}[t]
\centering
\small
\renewcommand{\arraystretch}{1.5}
\begin{tabularx}{\textwidth}{@{} l p{2.2cm} X p{2.5cm} @{}}
\toprule
\textbf{Query} & \textbf{Persona} & \textbf{System Generated Output (Representative Snippets)} & \textbf{Citations} \\ 
\midrule
\textbf{Q18} & Cynical \newline Journalist & "He reportedly lives in luxury, wearing a robe worth 160,000 RMB and driving an Audi luxury car... He has been dubbed the 'Shaolin CEO' for his commercial empire including Taobao shops and global trademarks." & \texttt{Xe2P8sYrT84} \newline \texttt{2-FA5-LZtyI} \\ 
\cmidrule{3-4}
 & & "Sensational reports alleged he was arrested at Pudong Airport while attempting to flee to Los Angeles with 7 mistresses and 21 children, though these were later labeled as 'fake'." & \texttt{2-FA5-LZtyI} \\ 
\midrule
\textbf{Q19} & Research \newline Analyst & "On July 27, 2025, the Shaolin Temple's official website released a situation report alleging Shi Yongxin was involved in occupational embezzlement and misappropriation of funds." & \texttt{2-FA5-LZtyI} \newline \texttt{BV1huSqB1EmS} \\ 
\cmidrule{3-4}
 & & "The Xinxiang People's Procuratorate approved the arrest of Shi Yongxin on November 16, 2025, for charges including non-state employee bribery and misappropriation of funds." & \texttt{BV1fmCQBWE2B} \newline \texttt{8N4n\_ArBp4A} \\ 
\bottomrule
\end{tabularx}
\caption{Comparative synthesis of the same video evidence under different persona lenses from our final submission. The system successfully bifurcates the narrative logic based on user background.}
\label{tab:case_study}
\end{table*}

As shown in Table~\ref{tab:case_study}, the system adopts a sensational and texture-rich tone for the Journalist (Q18), highlighting luxury assets and controversial rumors. In contrast, for the Analyst (Q19), it suppresses these distractors and constructs a rigorous, institutional timeline grounded in legal terminology and official reports. This divergence confirms that the A.I.R. agent acts as a strict cognitive filter over the SMC, ensuring that the generated content is not just factually correct, but contextually relevant to the specific user bias \cite{zheng2023judging}.

\section{Conclusion}

In this paper, we presented C2F-RAG, a fully training-free, cascaded Video RAG pipeline designed for large-scale multimodal augmented generation. By decoupling semantic pre-fetching from cognitive logical reasoning, our system effectively bridges the "semantic-logical gap" inherent in massive video collections. The architecture leverages BGE-M3 for high-recall coarse retrieval and an adapted A.I.R. agent for fine-grained cognitive reranking over the Serial Multimodal Context. 

Experimental results on the MultiVENT 2.0 corpus demonstrate that C2F-RAG achieves a state-of-the-art nDCG@10 of 0.848 and an Info F1-score of 0.801, significantly outperforming existing baselines. Furthermore, our system exhibits remarkable robustness, maintaining superior generation quality even when the search space scales from a few dozen to over one hundred thousand videos. Future work will focus on improving fine-grained temporal grounding recall in open-set environments. C2F-RAG provides a scalable and economically viable plug-and-play solution for complex, persona-constrained multimodal reasoning tasks.

\section*{Acknowledgment}
This work was supported by HUST Interdisciplinary Research Program under Grant No. 2025JCYJ077, the Ministry of Science and Technology of China under Grant No. 2025ZD0123800, and the KingSoft 2026 University-Industry Project. 


\bibliography{custom}

\appendix

\section{Official MAGMaR Challenge Results (Topic-Level)}
\label{app:official_results}

\begin{table*}[htbp]
\centering
\normalsize
\setlength{\tabcolsep}{18pt}
\begin{tabular}{lrrrr}
\toprule
Topic & \multicolumn{2}{c}{Info F1} & \multicolumn{2}{c}{Cite F1} \\
\cmidrule(lr){2-3} \cmidrule(lr){4-5}
 & P & R & P & R \\
\midrule
\textbf{Average} & 55.7 & 46.6 & 45.2 & 34.9 \\
\midrule
Liberation\_Day\_Tariffs\_q1 & 58.6 & 71.8 & 52.9 & 79.5 \\
Shi\_Yongxin\_Scandal\_q1 & 66.0 & 51.5 & 60.0 & 44.7 \\
2025\_Canadian\_federal\_election\_q2 & 47.8 & 69.4 & 41.3 & 55.6 \\
Blue\_Ghost\_Mission\_1\_q1 & 45.3 & 75.0 & 17.0 & 64.3 \\
Liberation\_Day\_Tariffs\_q2 & 50.0 & 59.0 & 42.0 & 59.0 \\
Shi\_Yongxin\_Scandal\_q2 & 56.0 & 51.5 & 66.0 & 51.5 \\
2025\_Alaskan\_Typhoon\_q1 & 71.4 & 42.9 & 7.1 & 0.0 \\
2025\_Myanmar\_earthquake\_q2 & 47.4 & 60.0 & 39.3 & 60.0 \\
2025\_Myanmar\_earthquake\_q1 & 40.0 & 73.3 & 40.0 & 40.0 \\
2025\_Alaskan\_Typhoon\_q2 & 68.1 & 41.3 & 4.3 & 0.0 \\
Blue\_Ghost\_Mission\_1\_q2 & 38.8 & 64.3 & 36.7 & 50.0 \\
Tropical\_Storm\_Wipha\_q2 & 62.3 & 31.7 & 66.7 & 27.4 \\
Palisades\_Fire\_q2 & 54.4 & 29.6 & 51.1 & 19.6 \\
Tropical\_Storm\_Wipha\_q1 & 77.8 & 24.6 & 75.0 & 20.3 \\
Central\_Texas\_Floods\_q1 & 59.0 & 26.7 & 56.4 & 15.6 \\
Nepal\_Youth\_Protests\_q1 & 59.6 & 23.5 & 60.9 & 14.7 \\
2025\_Canadian\_federal\_election\_q1 & 23.5 & 52.8 & 24.7 & 38.9 \\
Nepal\_Youth\_Protests\_q2 & 72.4 & 20.6 & 65.5 & 14.7 \\
Palisades\_Fire\_q1 & 60.3 & 16.9 & 52.4 & 7.9 \\
\bottomrule
\end{tabular}
\caption{Detailed topic-level evaluation results of our proposed C2F-RAG system for Information Generation (Info F1) and Temporal Grounding (Cite F1) under the Oracle setting using the CLUE framework.}
\label{tab:CLUE-reference-4-24-submission-Neofitos-Hofbeck}
\end{table*}

This section presents the comprehensive, topic-level evaluation results of our final system submission provided by the MAGMaR challenge organizers. While the main text reports the macro-averaged performance across all evaluation dimensions, Table~\ref{tab:CLUE-reference-4-24-submission-Neofitos-Hofbeck} offers a fine-grained breakdown of Information Generation (Info) and Temporal Grounding (Cite) metrics for individual query topics under the Oracle/Reference setting evaluated via the CLUE framework.

\section{Detailed Implementation of Prompt Sculpting}
\label{app:prompts}

\begin{figure*}[t]
\begin{lstlisting}[frame=single, basicstyle=\small\ttfamily, breaklines=true, title={Prompt Template A: Cognitive Filtering Engine (Fine Stage)}]
SYSTEM INSTRUCTION:
You are an advanced, hyper-logical document retrieval and filtering engine. Your sole objective is to analyze a provided JSON database of video contexts (including visual descriptions, OCR text, and ASR transcripts) and score each video's logical alignment with a specific user query and persona.

--- PART 1: THE RELEVANCE SCORING MATRIX ---
You must score every single video candidate against the user's `query` AND `persona`. A video might be highly relevant to the general topic, but completely irrelevant to the persona (a "Hard Negative"). Assign a relevance score from 0.00 to 1.00 using this strict rubric:

- [0.90 - 1.00] DIRECT & EXPLICIT ALIGNMENT: The video context contains explicit evidence that directly answers the core question of the query, perfectly matching the persona's needs.
- [0.75 - 0.89] STRONG CONTEXTUAL VALUE: Provides strong supporting information or secondary evidence related to the event.
- [0.50 - 0.74] TANGENTIAL / BACKGROUND RELEVANCE: Belongs to the correct event but provides little actionable intelligence for the specific persona.
- [0.30 - 0.49] WEAK / NOISY RELEVANCE: Barely related. Mentions the topic but focuses on different aspects.
- [0.00 - 0.29] IRRELEVANT (MUST BE EXCLUDED): Belongs to a different event or contains no useful information.

RULE: You must EXCLUDE any video with a score below 0.30. Sort the remaining videos in strictly descending order.

--- PART 2: STRICT OUTPUT SCHEMA ---
Output ONLY a raw JSON object. Do not use markdown formatting.
{
  "evaluations": [
    {
      "video_id": "<ID>",
      "reasoning": "<Justification based on the 5-tier rubric>",
      "relevance_score": 0.95
    }
  ]
}
\end{lstlisting}
\caption{Instructional skeleton for the Cognitive Filtering Agent. This stage focuses on logical pruning and score calibration.}
\label{fig:prompt_a}
\end{figure*}

The efficacy of C2F-RAG in navigating a 110,000-video corpus without prior fine-tuning relies heavily on the precision of our \textit{Prompt Sculpting} mechanism. While the main body of this paper conceptualizes these processes as the "A.I.R. Agent" and the "Serial Multimodal Context (SMC)", this appendix provides the actual instructional logic used to operationalize these concepts. We decouple the reasoning process into two distinct stages: a logical pruning stage and a persona-consistent synthesis stage.

\subsection{Fine Stage: Cognitive Filtering and Hard Negative Pruning}
As discussed in Section 2.3, the Fine Stage is designed to overcome the "semantic-logical gap" by simulating high-level human reasoning. The prompt for this stage (shown in Figure~\ref{fig:prompt_a}) is engineered to treat the SMC not merely as a text block, but as a multi-modal evidence database. 

A key design element is the 5-tier Relevance Scoring Matrix. By forcing the LLM to categorize videos into granular bins (e.g., distinguishing "Strong Contextual Value" from "Tangential Background"), we enable the system to implement the Logic-Gated Exponential Attenuation (LGEA) formula with high confidence. This stage explicitly penalizes "Hard Negatives"---videos that pass initial semantic filters due to keyword overlap but fail the persona's logical requirements.

\begin{figure*}[t]
\begin{lstlisting}[frame=single, basicstyle=\small\ttfamily, breaklines=true, title={Prompt Template B: Persona-Constrained Synthesis (Generation Stage)}]
SYSTEM INSTRUCTION:
You are an advanced, hyper-logical factual synthesis engine. You must act precisely as the persona described in the `background` field. Do not assume any external context or knowledge outside of the provided JSON database.

--- PART 1: LANGUAGE, BIAS, & PERSONA ADAPTATION ---
1. TARGET LANGUAGE OBLIGATION: You MUST write the generated text in the EXACT language specified by the user query (e.g., if "language": "nepali", you must use Nepali). JSON keys remain in English.
2. QUERY TYPE & BIAS CONTROL: 
   - If "biased": Embrace the subjective agenda or emotional angle of the persona.
   - If "unbiased": Remain strictly objective, neutral, and analytical.
3. Lexicon and Jargon: Filter the context through the eyes of the persona. Use domain-specific terminology (e.g., a statistician uses "variance, vote share").

--- PART 2: THE ZERO HALLUCINATION DIRECTIVE ---
You are strictly forbidden from injecting world knowledge. If the user asks "How many seats did the party win?" and the database only says "The party won," you must state: "The video data confirms the win, but specific seat counts are unavailable." NEVER invent OCR text or statistics.

--- PART 3: GENERATION & CITATIONS ---
1. Sentence-level Citations: Every single sentence you write MUST be supported by the database. You must append a `citations` array to EVERY sentence object.
2. Global References: Provide a single, flat, deduplicated list of every `video_id` cited.

--- PART 4: STRICT OUTPUT SCHEMA ---
Output exactly in the following JSON format. Do not use markdown blocks.
{
  "generation": {
    "responses": [
      {
        "text": "<Persona-aligned synthesized sentence in target language.>",
        "citations": ["<video_id>"]
      }
    ],
    "references": ["<video_id>"]
  }
}
\end{lstlisting}
\caption{Instructional skeleton for the Persona-Constrained Generator. This stage focuses on narrative synthesis, cross-lingual adaptability, and strict auditable grounding.}
\label{fig:prompt_b}
\end{figure*}

\subsection{Generation Stage: Persona Adherence and Grounding}
The final Generation Stage (Prompt B, Figure~\ref{fig:prompt_b}) shifts the focus from filtering to synthesis. The primary challenge here is maintaining the "Cognitive Lens" of the persona while ensuring zero-hallucination. 

To achieve this, we implement two critical safeguards. First, the Zero Hallucination Directive explicitly forbids the injection of external world knowledge, a common failure mode in LLMs when dealing with famous entities or events. Second, the Persona Adaptation instructions force a lexicon shift, ensuring that a "Statistician" uses technical jargon while a "Journalist" focuses on narrative texture. Furthermore, the prompt enforces a strict JSON schema for \textit{Deterministic Post-processing}, ensuring that the generated responses are perfectly formatted for automated evaluation and chunk-level temporal grounding.

\end{document}